# Assessing Pause Thresholds for empirical Translation Process Research

Devi Sri Bandaru, Michael Carl, Xinyue REN
CRITT, Kent State University

Translation production proceeds in the form of stretches of typing, interrupted by keystroke pauses. It is often assumed that fast typing reflects unchallenged/automated processes while long(er) typing pauses are indicative of translation problems, hurdles or difficulties. While there has been a long discussion concerning the determination of pause thresholds that separate automated from presumably reflective translation processes (O'Brien 2006; Alves and Vale 2009; Timarová et al. 2011; Dragsted and Carl 2013; Lacruz et al. 2014; Kumpulainen 2015; Heilmann and Neumann 2016), this paper compares three recent approaches for computing these pause thresholds.

## Methods and Metrics

We draw on a set of 478 from-scratch translation sessions from the CRITT TPR-DB that cover 10 different language pairs to assess two pause thresholds that have recently been proposed and a novel method. In contrast to previous approaches (e.g., Carl and Kay 2011; Couto-Vale 2017) these recent approaches take into account translator-specific typing speed based on the distinction between median within-word (WW) inter-keystroke intervals (IKIs[1]) and median between-word (BW) IKIs (Muñoz & Apfelthaler 2022, Miljanović et al., 2025, Carl et al 2025). WW-IKIs are shorter than BW-IKIs, presumably because automated typing processes are dominant when a typist is completing a word/thought that they have in mind (Crotti et al 2026). BW-IKIs are longer than WW-IKIs as they "are more likely to be linked to lexical retrieval" (Miljanović et al. 2025).

To compute WW and BW, we adopt the method suggested by Miljanović et al. (2025):

- WW: median delay between two successive alphanumeric keystrokes
- BW: median delay between an alphanumeric and a non-alphanumeric keystroke

Two cases can be distinguished for computing BWs: WS is a keystroke bigram that occurs at the end of a word (in alphabetical languages) when an alphanumeric keystroke (W, i.e., last character of a word) is followed by a non-alphanumeric keystroke (S, e.g., a blank space). SW occurs at the beginning of a new word, when a non-alphanumeric keystroke (S) is followed by an alphanumeric keystroke (W), first character of the next word. Surprisingly, the durations of these two types of BWs are quite different (see Figure 2): median WS-IKIs are much shorter, and their distribution more similar to WW, than median SW-IKIs, suggesting that the whitespace following a word is more tightly linked to the preceding word than to the following word. Accordingly, we base the BW computation on the longer SW-IKI. We compare three approaches:

[1] Translation sessions were collected with Translog-II (Carl 2012). Translog-II has a built-in key-logger (key&text logging for Chinese/Japanese) where an IKI is the lag of time between two successive key-down events. Note that Crotti et al (2026) define an IKI as "the lapse between the keyup of one key and the keydown of the next".

1. Muñoz & Apfelthaler (2022) distinguish between:
   - **RSP**: Respite: Unintentional short pause, defined as 2 × median WW-IKIs.
   - **TSP:** Task-Segment Pause: Intentional breaks, defined as 3 × median BW-IKIs.
2. Miljanović et al. (2025) suggest the following two measures:
   - **BW**: Processing effort: i.e., median BW-IKIs. (same as **SW**)
   - **MUD**: Micro-Unit Delimiter: defined as median WW-IKI + 2 × standard deviation
3. This paper:
   - **KUI**: Keystroke Interruption, defined as 2 × median WW-IKIs (same as **RSP**).
   - **PUB**: Production Unit Break, partially defined based on product data, as follows:

We suggest and evaluate a novel method for computing Production Unit Breaks (**PUBs**)[1]. The rationale for **PUB**s is grounded in the observation that text production proceeds in chunks of a few words (say 3 words) rather than word-by-word. That is, we are more likely to see an extended lexical retrieval (or reflection) pause every third word rather than every word. Accordingly, we take the IKI duration of the quantile for this ratio (text length/#chunks) as a threshold that separates automated from reflective processes. For instance, assuming that a word (in an English text) has on average 6 characters and a chunk length of 3 words, we take it that the upper 1/(6*3=18) IKI quantile—that is, ~5.5% of the longest IKIs—is related to intentional/reflective/lexical retrieval pausing, while the remaining 94.5% of the shorter IKIs are related to automated processes. Since the average length of words changes from text to text, also this threshold changes slightly.

Table 1 shows the Pearson (upper triangle) and Spearman correlations between the four measures (the measures **3GL** and **3GU** are discussed below). Note that there is a perfect correlation between **SW** and **TSP** as the latter is just a multiple of the former.

| | RSP | TSP | MUD | PUB | 3GU |
|---|---|---|---|---|---|
| **RSP** | 1 | *0.85* | *0.44* | *0.63* | *0.45* |
| **TSP** | 0.75 | 1 | *0.44* | *0.72* | *0.48* |
| **MUD** | 0.55 | 0.49 | 1 | *0.51* | *0.36* |
| **PUB** | 0.64 | 0.7 | 0.61 | 1 | *0.58* |
| **3GL** | 0.31 | 0.28 | 0.23 | 0.33 | 1 |
| **3GU** | 0.59 | 0.57 | 0.51 | 0.68 | 0.65 |

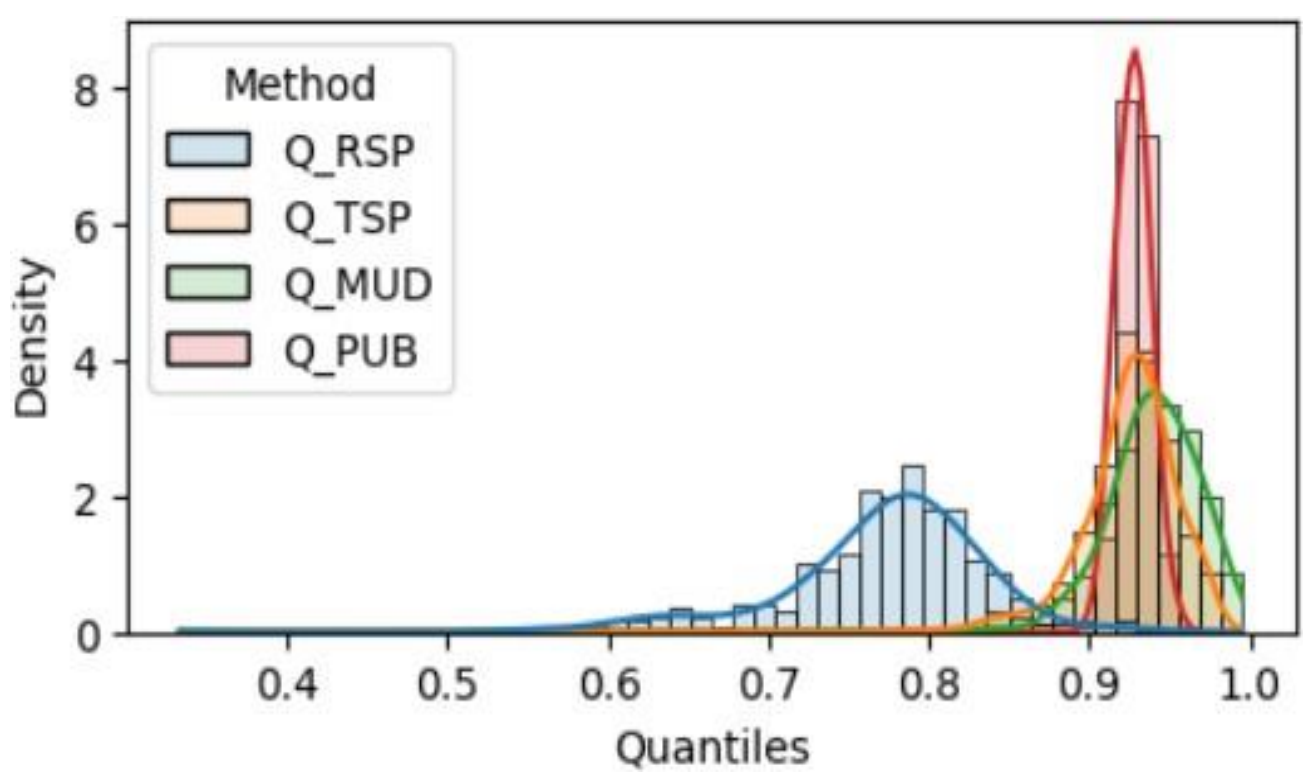

Table 1 (left): Correlation matrix of pause thresholds and GMM clusters; Upper triangle: Pearson correlation in italics; lower triangle: Spearman correlation. All Pearson correlation for **3GL** (not shown) are insignificant. All shown Spearman correlations are highly significant ($p < 0.001$). Figure 1 (right): distribution of quantiles for the four pause thresholds.

Figure 2 (left) plots the distribution of the two BW measures and, on the right, the 4 pause measures for the 478 translation sessions. Clearly, the distribution of SW-IKI is much flatter than the WS-IKI. **TSP** and **PUB** are strongly correlated, while **MUD** is much more stretched and less strongly correlated with the other measures.

[1] This version of **PUB** is different from the one described in (Carl et al 2025). It is the basis for the segmentation of Production and Translation Units in the TPR-DB 3.0, see https://critt-kent.github.io/TPR-DB-documentation/.

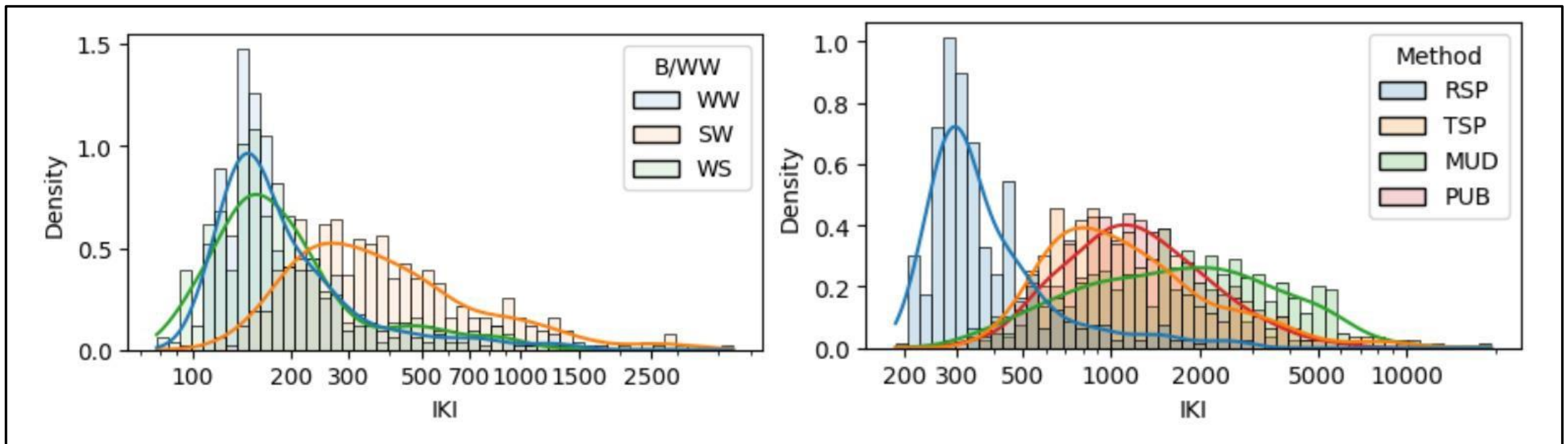

Figure 2: Left: distribution of within-word (WW) and two types of between-word BW-IKIs. Note that the distribution of WW is very similar to WS, suggesting that the separator following the word is still part of the word just typed, while SW-IKIs are indicative of different mental processing.
Right: four different pause thresholds: Note that RSPs and TSP are from the same distributions as WWs and SWs (left) multiplied by a factor of 2 and 3, respectively. Horizontal axes in both Figures are on a log scale.

## Validating Pausing Methods

While this analysis gives us a picture of how these methods distribute IKIs into two categories (presumably automated vs. reflective processes), it is unclear whether these thresholds/measures actually do their job, and which of them in the best way. To address this question, we investigated the distributions of IKIs per translation session with Gaussian Mixture Models (GMMs). GMMs identify clusters in the data without assuming theoretically imposed cutoffs (as our four methods do). If a (e.g. 2-component) GMM consistently separates IKIs (or fixation durations, see below) into two distinct distributions, this suggests that there may be two underlying cognitive processes generating these distinct timing patterns, potentially corresponding to automated versus reflective processing.

We applied the Akaike Information Criterion (AIC) to determine the optimal number of pause categories based on the empirical distribution of the data. The AIC analysis consistently favored a three-component model for IKIs—suggesting that the IKI distributions are actually composed of three independent processes—which are separated by two pause thresholds, providing strong statistical support for the theoretical framework proposed by Muñoz and Apfelthaler (2022). They posit three components:

1. **Automated Translation Routines:** fast-response component, short IKIs, corresponding to motor execution and fluent text production.
2. **Unintentional Halts (Respites):** micro-pauses where cognitive resources are momentarily drawn away from typing; yet no deep problem-solving involved.

3. **Intentional Stops (Reflective Pauses):** a slow-response component, probably corresponding to higher-order cognitive processing and decision-making.

The tripartite structure suggests two pause thresholds that temporarily separate the three processes. We label **3GL** the lower threshold and **3GU** for the upper threshold in Table 1. In our 'theory-driven' analysis we have so far one candidate (**RSPs**) for the lower threshold but three candidates (**TSP**, **MUD**, **PUB**) for the upper threshold which is supposed to separate unintentional halts from intentional stops. As shown in Table 1, there is virtually no correlation between the lower threshold (**3GL**) and the 'theoretical' metrics (not even **RSP**) but a moderate to strong positive correlation for the three upper threshold and the **3GU**.

This suggests all three analytic thresholds (**TSP**, **MUD**, **PUB**) are not merely arbitrary but align with a data-driven separation of processing modes; the **PUB** method, albeit, in the best way.

## Keystroke Pausing and Fixation Duration

We validate whether the two kinds of IKI thresholds identified in our keystroke analysis also correspond to different kinds of cognitive processing load as measured through fixation duration. It is generally assumed that longer fixations are indicative of higher cognitive load, which leads us to hypothesize that translators with longer average fixation durations might also show longer keystroke pauses.

We distinguished between visual attention allocated to the source text (ST, input processing), and the target text (TT, output monitoring). Furthermore, we separated the data into first fixations (FF) duration, indicative of processing effort for initial lexical access, and all fixations (AF), which captures the total dwelling time including re-reading and regressions.

We applied the GMM method to log-transformed fixation durations (LogDur), separately for FF and AF. In striking contrast to the tripartite structure observed in IKIs, the AIC analysis for ST and TT fixation durations consistently favored a two-component model (2G). This suggests that visual attention in translation operates more likely under a binary cognitive architecture, probably distinguishing between automated scanning (short fixations for navigation) and reflective processing (long fixations associated with active decision-making).

We then determined the point where the two fixation duration probability curves cross — that is, the duration at which a fixation is equally likely to belong to either the automated or reflexive component. This crossover point was calculated numerically using Brent's method, a standard root-finding algorithm. Fixations longer than this threshold are assumed to reflect more likely deliberate, conscious/reflective processing rather than automated reading behavior.

| Metric | 2G_ST_FF | 2G_ST_AF | 2G_TT_FF | PUB | RSP | TSP | MUD | 3GL | 3GU |
|---|---|---|---|---|---|---|---|---|---|
| **2G_ST_FF** | 1 | 0.89 | 0.75 | 0.35 | 0.36 | 0.31 | 0.18 | 0.14 | 0.28 |
| **2G_ST_AF** | 0.89 | 1 | 0.81 | 0.35 | 0.4 | 0.37 | 0.2 | 0.13 | 0.29 |
| **2G_TT_FF** | 0.75 | 0.81 | 1 | 0.28 | 0.39 | 0.37 | 0.23 | 0.12 | 0.28 |
| **2G_TT_AF** | 0.81 | 0.89 | 0.9 | 0.31 | 0.43 | 0.4 | 0.21 | 0.12 | 0.28 |

**Table 2**: Spearman correlations for fixation and pause thresholds. Rows: fixation 2-component GMM thresholds for ST/TT and FF/AF. The correlations with **3GL** (GMM lower keystroke threshold) are very weak but significant $p < 0.05$. All other correlations are highly significant $p < 0.01$.

We compute this **2G** threshold for the ST and TT and for FF and AF and correlate the threshold values with keystroke pause metrics —**PUB**, **3GL**, **3GU**, **TSP**, **RSP**, and **MUD** — over all 478 translation sessions. As shown in Table 2, the four **2G** thresholds strongly correlate with each other, indicating that these measures are not independent. The analysis also reveals a consistent but moderate correlation between most production pauses and fixation durations. Specifically, the **PUB**, **RSP**, **TSP** and **3GU** aligns more closely with AF metrics (for both ST and TT) than with FF, suggesting that intentional pauses are more frequently associated with late gaze measures (re-reading, regressions, verification) rather than initial lexical access.

Table 2 provides the Spearman correlation, but Pearson correlation consistently provided lower values (not shown). The divergence between these two correlations indicates that the relationship between gaze durations and keystroke pauses is primarily monotonic rather than strictly linear; suggesting that visual (cognitive) load increases as keystroke pauses increase. Notably, the **RSP** and **TSP** thresholds exhibit the strongest monotonic (Spearman) correlations with TT fixations, whereas the lower keystroke threshold **3GL,** shows very weak correlations across all gaze measures. This weak correlation in Tables 1 and 2 suggests that **3GL** likely captures motor execution noise rather than cognitively driven pauses. However, the significant correlation across the higher thresholds suggests that the thresholding approach captures a robust and reliable signal of shifts between automatic and reflective processing modes.

## Data Availability:

The data and the python scripts for this study can be downloaded from: https://github.com/Critt-Kent/Key-Gaze-Correlation